\def\year{2021}\relax

\documentclass[letterpaper]{article}

\usepackage{aaai21} 
\usepackage{times} 
\usepackage{helvet} 
\usepackage{courier} 
\usepackage[hyphens]{url} 
\usepackage{graphicx} 
\usepackage{graphicx} 
\usepackage{natbib}
\usepackage{caption}
\usepackage{xr}
\usepackage[utf8]{inputenc} 
\usepackage[T1]{fontenc}    

\usepackage{booktabs}       
\usepackage{amsfonts}       
\usepackage{nicefrac}       
\usepackage{microtype}      
\usepackage{multirow}				
\usepackage{subfig}					
\usepackage{cancel}					
\usepackage{amsmath}				
\usepackage{placeins}				
\usepackage{mathtools}			
\usepackage{resizegather}   
\usepackage[pdf]{graphviz}	

\usepackage{algorithm}
\usepackage{algpseudocode}

\usepackage{amsthm}

\title{Identification of Latent Variables From Graphical Model Residuals}
\author{Boris Hayete \thanks{equal contribution}, Fred Gruber \footnotemark[1], Anna Decker, Raymond Yan}

\newtheorem{theorem}{Theorem}

\begin{document}
\maketitle

\begin{abstract}
Graph-based causal discovery methods aim to capture conditional independencies consistent with the observed data and differentiate causal relationships from indirect or induced ones.  Successful construction of graphical models of data depends on the assumption of causal sufficiency: that is, that all confounding variables are measured. When this assumption is not met, learned graphical structures may become arbitrarily incorrect and effects implied by such models may be wrongly attributed, carry the wrong magnitude, or mis-represent direction of correlation.  Wide application of graphical models to increasingly less curated "big data" draws renewed attention to the unobserved confounder problem.  

We present a novel method that aims to control for the latent space when estimating a DAG by iteratively deriving proxies for the latent space from the residuals of the inferred model.  Under mild assumptions, our method improves structural inference of Gaussian graphical models and enhances identifiability of the causal effect. In addition, when the model is being used to predict outcomes, it un-confounds the coefficients on the parents of the outcomes and leads to improved predictive performance when out-of-sample regime is very different from the training data.  We show that any improvement of prediction of an outcome is intrinsically capped and cannot rise beyond a certain limit as compared to the confounded model.  We extend our methodology beyond GGMs to ordinal variables and nonlinear cases.  Our R package provides both PCA and autoencoder implementations of the methodology, suitable for GGMs with some guarantees and for better performance in general cases but without such guarantees. 
\end{abstract}

\section{Introduction}
\label{introduction}

Construction of graphical models (GMs) pursues two related objectives:
accurate inference of conditional independencies (\textit{causal
  discovery}), and construction of models for outcomes of interest
with the purpose of estimating the average causal effect (ACE) with
respect to interventions \cite{pearl_causality:_2000,hernan_estimating_2006} (\textit{causal inference}).  These two distinct goals mandate that certain identifiability conditions for both types of tasks be met.  Of particular interest to us is the condition of \textit{causal sufficiency} (\cite{spirtes_causation_1993}) - namely, that all of the relevant confounders have been observed.  When this condition is not met, accurate GMs and identifiability of the causal effects are hard to infer.

Intuitively, the presence of unobserved confounding leads to violations of conditional independence among the affected variables downstream from any latent confounder.  Score-based methods for GM construction aim to minimize unexplained variance for all variables in the network by accounting for conditional independencies in the data \cite{pearl_causality:_2000,friedman_being_2013}.  Unobservability of a causally important variable will induce dependencies among its descendants that are conditionally independent of each other given the latent variable.  This gives rise to inferred connectivity that is excessive compared to the true network \cite{elidan_discovering_2001}.  Further, since none of the observed descendants are perfect correlates of the unobserved ancestor, some of the information from the ancestor will "pollute" model residuals and give rise to correlation patterns in the residual space.   

Hitherto, the causal inference literature has largely focused on addressing the subset of problems where the ACE can be estimated reliably in the absence of this guarantee, such as when conditional exchangeability holds \cite{hernan_estimating_2006}.  In causal discovery, certain constraint-based methods like the Fast
Causal Algorithm (FCI) \cite{spirtesCausalInferencePresence1995} can
deal with general latent structures but do not scale well to large
problems. While faster variants of FCI like RFCI \cite{colombo_learning_2012}
and  FCI+ \cite{claassenLearningSparseCausal2013} exist, score-based
approaches tend to perform better than purely constraint-based methods. This
has been demonstrated in problems without latent confounders
\cite{nandy2018high} and has been suggested that this is because
constraint-based methods are sensitive to early mistakes which
progagate into more errors later on while in score-based approaches
the mistakes are localized and do not affect the score of graphs later
on \cite{bernstein2020ordering}. It is likely that these limitations also
affect FCI-type algorithms. Furthermore, while FCI can deal with
general latent structures, in certain scenarios where a few latent confounders
drive  many observed variables  most observed variables are
conditionally dependent which implies dense maximal ancestral graphs
where very few edges can be oriented
\cite{frotRobustCausalStructure2017} (see supplementary Figure \ref{sup-fig_rfci_ex1} for
an illustration). Finally, another disadvantage of
constraint based methods is that they generate a single model and do
not provide a score summarizing how likely the network is given the
data while score based methods can generate this type of scores by using Bayesian or ensemble approaches \cite{jabbariDiscoveryCausalModels2017}. Relatively computationally efficient score-based methods have been proposed for inferring latent variables affecting GMs by the means of expectation maximization (EM) (as far back as \cite{friedman1997learning,friedman1998bayesian}).  However, for a large enough network, local gradients do not provide a reliable guide, nor do they address the cardinality of the latent space.  Methods for using near-cliques for detection of latent variables in directed acyclic graphs (DAGs) with Gaussian graphical models (GGMs) have been proposed that address both problems by analyzing near-cliques in DAGs \cite{elidan_discovering_2001,silva_learning_2006}.  A method related to ours has been proposed for calculating latent variables in a greedy fashion in linear and "invertible continuous" networks, and relating such estimates to observed data to speed up structure search and enable discovery of hidden variables \cite{elidan_ideal_2007}. However, with a clique-driven approach it is impossible to tell whether any cliques have shared parents and, importantly, whether any signal remained to be modeled, resulting in score-based testing rejecting proposed "ideal parents". Additionally, Wang and Blei (2019) introduced the deconfounder approach to detecting and adjusting for latent confounders of causal effects in the presence of multiple causes but in the context of a fixed DAG under relatively strict conditions \cite{wang_deconfounder_2019,wangMultipleCausesCausal2019}. 

We show that there exist circumstances when causal sufficiency can be asymptotically achieved and exchangeability ensured even when the causal drivers of outcome are confounded by a number of latent variables.  This can be done when confounding is \textbf{pleiotropic}, that is,  when the latent variable affects a "large enough" number of variables, some driving an outcome of interest and others not (we tie this to the expansion property defined in \cite{anandkumar_learning_2013}).  Notably, this objective cannot be achieved when confounding affects only the variables of interest and their causal parents (\cite{damour_multi-cause_2019}).  Our approach is provably optimal given a proposed graph. This leads to the iterative EM-like optimization approach over structure and latent space proposed below.

The outline of this paper is as follows: In the first two sections, we discuss the process of diagnosting latent confounding in the special case of GGMs where closed-form local solutions exist. In the third section, we derive the improvement cap for predictive performance of the graph thus learned. In the fourth section, we discuss implementation, and in the fifth demonstrate the approach on simulated and real data. Finally, we propose some extensions to the approach for more complex graphical model structures and functional forms. 

\section{Background And Notation}
\label{Background}

Consider a factorized joint probability distribution over a set of observed and latent variables Table \ref{tab:notation} summarizes the notation to be used to distinguish variables, their relevant partitions , and parameters.

\begin{table}[h]
\centering
\scalebox{0.85}{
\begin{tabular}{ c|c|c } 
 Symbol & Meaning & Indexing \\ 
 \hline
 $S$ & samples & $S_i, i \in \{1, \dots, s\}$ \\ 
 $V$ & observed predictor variables & $V_j, j \in \{1, \dots, v\}$ \\
 $U$ & unobserved predictor variables & $U_l, l \in \{1, \dots, u\}$ \\
 $O$ & outcomes (sinks) & $O_k, k \in \{1, \dots, o\}$ \\
 $D$ & $\{V, O\}$ - observable data matrix & \\
 $\bar{D}$ & estimate of variables in D & \\
 & from their parents & \\
 $D_u$ & $\{V, O, U\}$ - implied data matrix& \\
 $\theta$ & parameters & $\theta_i, i \in \{1, \dots, t\}$ \\
 $Pa^N$ & parents of variable $N$ & $Pa^N_i, i \in \{1, \dots, p\}$\\
 $C^N$ & children of variable $N$ & $C^N_i, i \in \{1, \dots, c\}$\\
 $G$ & graph over $D$ &\\
 $G_u$ & graph over $D_u$& \\
 $R$ & residuals - matrix matching $D$ &\\
\end{tabular}}
\caption{Notation}
\label{tab:notation}
\end{table}

Assume that the joint distribution $D_{u}$ over the full data or $D$ over the observed data is factorized as a directed acyclic graph, $G_{u}$ or $G$.  We will consider individual conditional probabilities describing nodes and their parents, $P(V | Pa^{V}, \theta)$, where $\theta$ refers to the parameters linking  the parents of $V$ to $V$.  $\hat{\theta}$ will refer to an estimate of these parameters.  We will further assume that $G$ is constructed subject to regularization and using unbiased estimators for $P(V | Pa^{V}, \hat{\theta})$.  We will further assume that $D_u$ plus any given constraints are sufficient to infer the true graph up to Markov equivalence.  For convenience, we'll focus on the actual true graph's parameters, so that, using unbiased estimators, $E[\hat{\theta_m}|D_u] = \theta_m, \forall m$.

Mirroring $D$ (or $D_u$), we will define a matrix $R$ of the same dimensions ($s \times (v + o)$) that captures the residuals of modeling every variable $N \in \{V, O, (U)\}$ via $G$ (or $G_u$).  Note that $R_u$ - the residuals matrix that contains residuals of $U$ - doesn't make sense and isn't defined.  In the linear case, these would be regular linear model residuals, but more generally we will consider probability scale residuals (PSR, \cite{shepherd_probability-scale_2016}).  That is, we define $R[i, j] = PSR(P(V_j | Pa^{V_j}, \hat{\theta}_j) | D[i, j])$, the residuals of $V_j$ given its graph parents.  The use of probability-scale residuals allows us to define $R$ for all ordinal variable types, up to rank-equivalence.

\section{Diagnosing Latent Confounding} 

Here, we consider the special case of GGMs where our approach for
determining the existence of latent confounders can be written down in a closed form. 

Recall that, for some $V_j \in \{V, U, O\}$, $P^{V_j}_k$ denotes the $k$th parent of $V_j$.  For GGMs, we can write down a fragment of any DAG G as a linear equation where a child node is parameterized as a linear function of its parents:
\begin{equation}
\begin{align*}
V_j = \beta_{j0} + \beta_{j1} P^{V_j}_1 + \dots + \beta_{jp} P^{V_j}_p + \xi_j,\\\xi_j \sim \mathcal{N}(0, \sigma_j).
\end{align*}
\end{equation}	

\begin{figure}[h]
	\centering
	\digraph[scale=0.5]{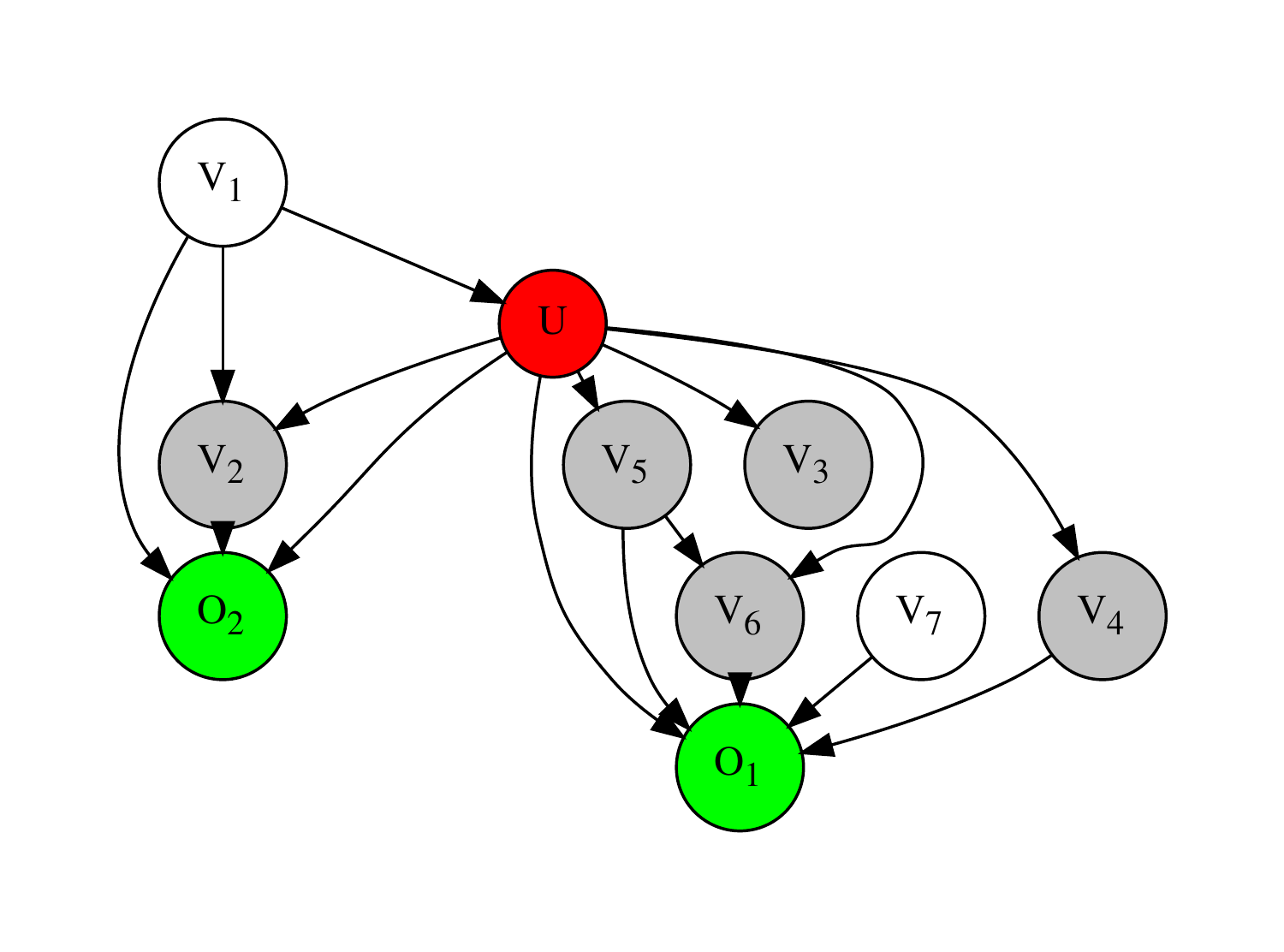}{
			graph[ranksep=0.1];
      node [shape=circle, style="filled"];
      U [fillcolor=red];
      V1 [fillcolor=white, label=<V<SUB>1</SUB>>];
      V2 [fillcolor=gray, label=<V<SUB>2</SUB>>];
      V3 [fillcolor=gray, label=<V<SUB>3</SUB>>];
      V4 [fillcolor=gray, label=<V<SUB>4</SUB>>];
      V5 [fillcolor=gray, label=<V<SUB>5</SUB>>];
      V6 [fillcolor=gray, label=<V<SUB>6</SUB>>];
      V7 [fillcolor=white, label=<V<SUB>7</SUB>>];
      O1 [fillcolor=green, label=<O<SUB>1</SUB>>];
      O2 [fillcolor=green, label=<O<SUB>2</SUB>>];
      V1 -> U; U -> V2; U -> V3; U-> V4; U-> V5; U -> V6; V1 -> O2; V2 -> O2; V4-> O1; V5 -> O1; V6 -> O1; U -> O2; U -> O1; V1 -> V2; V5 -> V6; V7 -> O1
	}
	\caption{Graph $G_u$.  $U$ influences the outcomes $O$, and a number of predictors $V$, confounding many of the $V_j \rightarrow O_k$ relationships.  Gray nodes are affected by $U$.}
	\label{fig:sampleGraph}
	\digraph[scale=0.5]{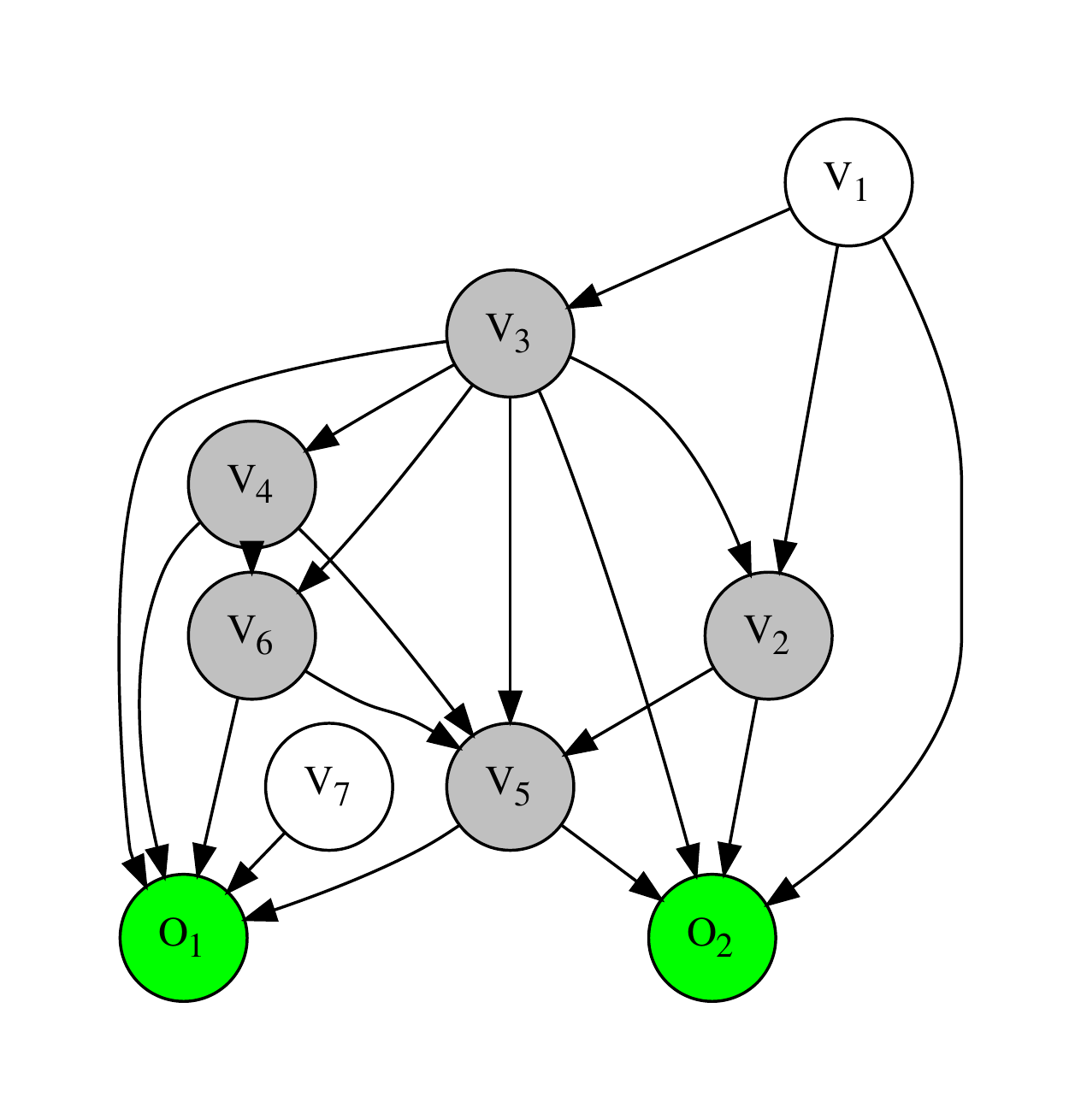}{
			graph[ranksep=0.1];
      node [shape=circle, style="filled"];
      V1 [fillcolor=white, label=<V<SUB>1</SUB>>];
      V2 [fillcolor=gray, label=<V<SUB>2</SUB>>];
      V3 [fillcolor=gray, label=<V<SUB>3</SUB>>];
      V4 [fillcolor=gray, label=<V<SUB>4</SUB>>];
      V5 [fillcolor=gray, label=<V<SUB>5</SUB>>];
      V6 [fillcolor=gray, label=<V<SUB>6</SUB>>];
      V7 [fillcolor=white, label=<V<SUB>7</SUB>>];
      O1 [fillcolor=green, label=<O<SUB>1</SUB>>];
      O2 [fillcolor=green, label=<O<SUB>2</SUB>>];
      V1 -> V3; V3 -> V2; V3 -> V4; V3 -> V6; V3 -> V5; V4 -> V6; V6 -> V5; V1 -> O2; V2 -> O2; V4-> O1; V5 -> O1; V6 -> O1; V7 -> O1; V3 -> O2; V4 -> V5; V5 -> O2; V1 -> V2; V2 -> V5; V3 -> O1
	}
	\caption{Graph $G$.  With $U$ latent, the graph adjusts, introducing spurious edges.}
	\label{fig:sampleGraphOnObservables}
\end{figure}

For example, consider $O_1$ in Figure \ref{fig:sampleGraph}.  We can write:
\begin{equation}
\begin{split}
O_1 = \beta_0 + \beta_6V_6 + \beta_5V_5 + \beta_4V_4 + \beta_7V_7 + \beta_uU + \xi_1,\\ \xi_1 \sim \mathcal{N}(0, \sigma_{O_1}).
\end{split}
\end{equation}

For any variable N that has parents in $G_u$, we can group variables in $P^N$ into three subsets: $X_U \in \{P^U, C^U\}$, $X_{\cancel{U}} \notin \{P^U, C^U\}$, and the set $U$ itself, and write down the following general form using matrix notation:
\begin{equation}
\begin{split}
N = \beta_{N0} + B_U X_U + B_{\cancel{U}} X_{\cancel{U}} + \beta_U U + \xi_N,\\\xi_N \sim \mathcal{N}(0, \sigma_{N}).
\end{split}
\label{eq:linearForm}
\end{equation}

Explicit dependence of $N$ on $U$ happens when $\beta_U \neq 0$.  

Note that if we deleted $U$ and its edges from $G_u$ without relearning the graph, Equation \ref{eq:linearForm} from $G_u$ would read:
\begin{equation}
N = \beta_{N0} + B_U X_U + B_{\cancel{U}} X_{\cancel{U}} + R_N + \xi_N 
\label{eq:linearFormNoU}
\end{equation}
where the residual term $R_N$ is simply equal to the direct
contribution of $U$ to $N$.

Now consider $G$,  the graph learned over the variables $\{V, O\}$ excluding the latent space $U$. The network $G$ would have to adjust to the missingness of $U$ (e.g., Figure \ref{fig:sampleGraphOnObservables} vs Figure \ref{fig:sampleGraph}).  As a result, $R_N$ will be partially substituted by other variables in $\{P^U, C^U\}$.  Still, unless $U$ is completely explained by $\{P^U, C^U\}$ (as described in \cite{damour_multi-cause_2019}) and in the absence of regularization (when a high enough number of covariates may lead to such collinearity), $R_N$ will not fully disappear in $G$.  Hence, even after partially explaining the contribution of $U$ to $N$ by some of the parents of $N$ in $G$, 
\begin{equation}
R_N = \beta_0 + \beta_1 U + \xi_N.
\label{eq:residualColumn}
\end{equation}

\begin{center}
\begin{table}[ht]
\centering
\scalebox{1}{
\begin{tabular}{c|c|c|c|c|}
&\multicolumn{4}{c}{Variables}\\
\hline
\multirow{3}{*}{\rotatebox[origin=c]{90}{Samples  }}
&$V_{11}$&$V_{12}$&\dots&$V_{1v}$\\
&\dots&\dots&\dots&\dots\\
&$V_{s1}$&$V_{s2}$&\dots&$V_{sv}$\\
\end{tabular}}
\qquad
\qquad
\qquad
\qquad
\scalebox{1}{
\begin{tabular}{c|c|c|c|c|}
&\multicolumn{4}{c}{Residuals}\\
\hline
\multirow{3}{*}{\rotatebox[origin=c]{90}{Samples  }}
&$R_{11}$&$R_{12}$&\dots&$R_{1v}$\\
&\dots&\dots&\dots&\dots\\
&$R_{s1}$&$R_{s2}$&\dots&$R_{sv}$\\
\end{tabular}}
\caption{The training data frame (Variables) implies a matching data frame (Residuals) once the joint distribution of all variables is specified via a graph and its parameterization}
\label{tab:MatchingResiduals}
\end{table}
\end{center}

Therefore, the columns in the residuals table corresponding to $G$ (Table \ref{tab:MatchingResiduals}) that represent the parents and children of $U$ will contain residuals collinear with $U$:
\begin{equation}
\begin{split}
R_1 = \beta_{10} &+ \beta_{11} U + \xi_1\\
& \hdots\\
R_k = \beta_{k0} &+ \beta_{k1} U + \xi_k.
\end{split}
\end{equation}

Rearranging and combining,
\begin{equation}
U = \beta_i^{*} R_1 + \dots + \beta_k^{*} R_k + \xi = B R + \xi.
\label{eq:resPCA}
\end{equation}

Equation \ref{eq:resPCA} tells us that, for graphical Gaussian models, components of $U$ are obtainable from linear combinations of residuals, or principal components (PCs) of the residual table $R$.  In other words, $U$ is identifiable by principal component analysis (PCA).  Whether the residuals needed for this identification exist depends on the \textit{expansion property} as defined in \cite{anandkumar_learning_2013}.

\begin{theorem}[Inferrability]
\label{thm:inferrability}
Posit that we know the true DAG $G$ over $D$, but not $U = G_u \setminus G$, the latent space's edges to observed variables.  Assume that $L$ is orthogonal, with no variable in $D$ a parent of any variable in $U$. Define as $R_C = C^{U,G} - \bar{C}^{U,G}$ the residuals of all children of $U$ that also have parents in $G$.  If relationships described by $G$ are linear, we can infer $U$ from $R_C$ up to sign and scale.  

\begin{proof}
For proof, refer to supplement, theorem \ref{sup-thm:inferrability}.
\end{proof}
\end{theorem}

Theorem \ref{thm:inferrability} does require G to be known, which will give rise to an iterative algorithm later on.  However, from it we see that focusing on residuals of a graph permits identification of the latent space in some circumstances.  Supplementary theorem \ref{sup-thm:deconfounding} relaxes the need for orthogonality when latent space's structure is irrelevant but controlling for it is important.

Note that this approach is similar, in the linear case, to that in \cite{elidan_ideal_2007} except insofar as we show the principal components of the entire residual matrix to be optimal for the discovery of the whole latent space of a given DAG, and thus couple structure inference and latent space discovery via EM (\ref{alg:latentEM}).

\section{Capping improvement}

The most important utility of causal modeling is to identify drivers as well as drivers of drivers of outcomes of interest.  These drivers of drivers may have practical importance. For example, in the development of a drug, direct predictors of an outcome (e.g. $V_6$ pointing to $O_{1}$ in Figure \ref{fig:sampleGraph}) may not be viable targets but upstream variables (e.g. $V_{5}$) may in fact be be promising targets and the direct parents of the outcome mediate the effect of these potential targets. In the presence of latent confounding, the identifiability of direct causal effects of outcomes is also jeopardized \cite{hernan_estimating_2006}.  For example, in Figure \ref{fig:sampleGraphOnObservables} misisng $U$ induces apparent correlation among $V_3$, $V_4$, $V_5$, and $V_6$ even though some of these variables are conditionally independent given $U$.  

The extent of the problem of unmeasured confounding can be quantified in more detail.  Suppose we model $O_1$ without controlling for $U$ (\ref{fig:sampleGraphOnObservables}): $$O_1 = \beta_0 + \beta_3 V_3 + \beta_4 V_4 + \beta_5 V_5 + \beta_6 V_6 + \dots.$$  Setting the coefficient of determination for the model $$V_3 = \alpha_0 + \alpha_4 V_4 + \alpha_5 V_5 + \alpha_6 V_6 + \dots.$$ equal to $\rho_3^2$.  Then the estimated variance of $\beta_3$ in the presence of collinearity can be related to the variance when collinearity is absent via the following formula \cite{rawlings_applied_1998}:
\begin{equation}
var(\bar{\beta}_3) = var(\beta_3) \frac{1}{1-\rho_3^2} \propto \frac{1}{1-\rho_3^2}.
\label{eq:VIF}
\end{equation}
Formula \ref{eq:VIF} describes the \textit{variance inflation factor} (VIF) of $\beta_3$.  Note that $\lim_{\rho \to 1} \frac{1}{1-\rho^2} = \infty$, so even mild collinearity induced by latent variables can severely distort coefficient values and signs, and thus estimation of ACE.  Our approach reduces the VIFs of coefficients related to outcomes and thus make all \textit{causal} statements relating to outcomes, such as calculation of ACE, more reliable, by controlling for $\bar{U}$ - the estimate of $U$ - in the network,
\begin{equation}
\lim_{(U - \bar{U})\to0} var(\bar{\beta}_i) = var(\beta_i).
\label{eq:vifImprovement}
\end{equation}

Consider an output $O_j$.  While it is difficult to describe the limit of error on the coefficients of the drivers of $O_j$, it is straightforward to put a ceiling on the improvement in the likelihood obtainable from modeling $U$ and approximating $G_u$ with $G_{\bar{u}}$.  Suppose we eventually model $U$ as a linear combination of a set of variables $X \subset V$, and denote by $X \setminus W$ the set difference: members of $X$ not in $W$.  Then for any outcome $O_i$ predicted by a set of variables $W$ in the graph $G$ and in truth predicted by the set $Z + U$, we can contrast three expressions (from $G$ and $G_{\bar{U}}$ respectively):
\begin{equation}
\begin{split}
O_i = \beta_{i0} + B_W W (a) + \xi_i \qquad (a)\\
O_i = \beta_{i0} + B_W W + B_{X \setminus W} (X \setminus W)
 + \xi_i \qquad (b)\\
O_i^U = \beta_{i0}^U + B_Z^U Z + B^U U + \xi_i \qquad (c).
\end{split}
\end{equation}
Model (a) is the model that was actually accepted, subject to regularization, in G.  Model (b) is the "complete" model of $O_i$ that controls for $U$ non-parsimoniously by controlling for all variables affected by $U$ and not originally in the model.  The third model, (c), is the ideal parsimonious model when U is known.  We can compare the quality of these models via the Bayesian Score, and the full score, which can approximated by the Bayesian Information Criterion (BIC) in large samples \cite{koller_probabilistic_2009}.  We assume that model (c) would have the lowest BIC (being the best model), and model (a) would be slightly better, since we know that the set of variables $X \setminus W$ didn't make it into the first equation subject to regularization by BIC.  Assuming $n$ samples,

\begin{align}
b_{a} &= BIC(O_{i} = \beta_{i0} + B_W W + \xi_i)  \\
b_{b} &= BIC(O_i = \beta_{i0} + B_W W  \\
               &  + B_{X \setminus W} (X \setminus W) + \xi_i) \nonumber \\
&  = b_c + |X\setminus W| log(n) \nonumber \\
b_{c} &= BIC(O_i^U = \beta_{i0}^U + B_Z^U Z + B^U U + \xi_i)
\end{align}

The "complete model" - model (b) - includes all of the true predictors of $O_j$.  Therefore its score will be the same as that of the true model - model (c) - plus the BIC penalty, $log(n)$, for each extra term, minus the cost of having $U$ in the true model (that is, the cardinality of the relevant part of the latent space).  We know that the extra information carried by this model was not big enough to improve upon model (a), that is $b_a < b_c + k \log(n)$ for some $k$.  Rearranging:
\begin{equation}
    b_c - b_a > -k\log(n).
    \label{eq:ceilingTheorem}
\end{equation}
Any improvement in $G_{\bar{U}}$ owing to modeling of $\bar{U}$ cannot, therefore, exceed $k\log(n)$ logs, where $k = |X \setminus W| - |U|$: the information contained in the "complete" model is smaller than its cost.

Although the available improvement in predictive power is also capped in some way, it is still important to aim for that limit.  The reason is, correct inference of causality, especially in the presence of latent variables, is the only way to ensure transportability of models in real-world (heterogeneous-data) applications (see, e.g., \cite{bareinboim_causal_2016}).

This approach is a generalization of work presented in \cite{anandkumar_learning_2013}, where the authors show that, under some assumptions. the latent space can be learned exactly, which is also related to the deconfounder approach described by \cite{wang_deconfounder_2019}.  However, our approach does not require that the observables be conditionally independent given the latent space and instead \textit{generate} such independence by the use of causal network's residuals, which are conditionally independent of each other \textit{given the graph and the latent space}.  However, since the network among the observables is undefined in the beginning, the structure of the observable network must be learned at the same time as the structure of the latent space, which leads us to the iterative/variational bayes approach presented in Algorithm \ref{alg:latentEM}.  Lastly, the use of the entirety of the residual space is different from the work described in \cite{elidan_ideal_2007}, where local residuals are pursued with the goal to accelerate structure learning while simultaneously discovering the latent space.

\section{Implementation}
Algorithm \ref{alg:latentEM} below describes our approach to learning the latent space and can be viewed as a type of an expectation-maximization algorithm.

How do we learn $\bar{U} = f(R_{\bar{U}})$?  In the linear case, we can use PCA, as described above, and in the non-linear case, we can use non-linear PCA, autoencoders, or other methods, as alluded to above.  However, the linear case provides a useful constraint on dimensionality, and this constraint can be derived quickly.  A useful notion of the ceiling constraint on the linear latent space dimensionality can be found in \cite{gavish_optimal_2014}.  From a practical standpoint, the dimensionality can be set even tighter, and we use a previously described heuristic approach\cite{buja_remarks_1992}.

\begin{algorithm}
 \caption{Learning $\bar{U}$ from structure residuals via EM}
 \label{alg:latentEM}
\begin{algorithmic}
 \State \textbf{Data:} The set of observed variables $\{V, O\}$
 \State \textbf{Result:} Graph $G_{\bar{U}}(V, O, \bar{U})$
 \State Construct $G = G(V, O)$\;
 \State Compute $S_0 = BIC_{G}$\;
 \State Estimate $\bar{U} = f(R)$\;
 \State Construct $G_{\bar{U}} = G(V, O, \bar{U})$\;
 \State Compute $S_{\bar{U}} = BIC(G_{\bar{U}})$\;
 \While{$S_{\bar{U}} - S_0 > \epsilon$} 
  \State Set $S_0 = S_{\bar{U}}$\;
  \State Calculate $R_{\bar{U}}$:\;
  \State Set $\bar{U}$ to arbitrary constant values\;
  \ForAll{child node $C \in G_{\bar{U}}, C \notin \bar{U}$}
  	\State Set parents to training data\;
  	\State $\bar{C} = C|parents(C)$\;
  	\State Set $R_C = PSR(\bar{C}, C)$\;
  \EndFor
  \State Estimate $\bar{U} = f(R_{\bar{U}})$\;
  \State Construct $G_{\bar{U}} = G(V, O, \bar{U})$\;
  \State Compute $S_{\bar{U}} = BIC(G_{\bar{U}})$\;
 \EndWhile
\end{algorithmic}
\end{algorithm}

\subsection{Nonlinear Extension: Autoencoder}

All results described above refer to Gaussian or at most rank-monotonic relationships, and perhaps extend to linear models with interactions, when interactions can be seen as "synthetic features". Real-world data, however, often does not behave this way requiring an approach that might generalize beyond monotonic relationships.  Therefore, we pursued latent space discovery using autoencoders.

We first assess the cardinality of the latent space (number of nodes in the coding layer) using the linear approach (PCA), and take this number as a useful "ceiling" for the dimensionality of the non-linear latent space, on the assumption that non-linear features are more compact and require lower dimensionality if discovered.  We then constructed an autoencoder with 4 hidden layers where the maximum number of nodes in the hidden layers was capped at $\min (100, \textrm{number of variables})$, the cap being dictated by practical considerations.

The autoencoder was implemented using Keras with Tensorflow backend and called within R using the Reticulate package \cite{reticulate_2020}. The encoders and decoder were kept symmetric and in order improve the stability we used tied weights \cite{pca_ae}. In addition, the coding layer had additional properties borrowed from PCA including a kernel regularizer promoting orthogonality between weights and an activity regularizer to promote uncorrelated encoded features (see \cite{ranjan_build_2019} for details on implementation and justification). This last property is of particular interest in our application since ideally every dimension in our latent space should be associated with a different latent variable. The hidden layers used a sigmoidal activation except for the output layer which had a linear activation. All layers had batch normalization \cite{ioffe2015batch} (see supplementary Figure \ref{sup-fig_ae} for a diagram of the architecture).

For more details see the implementation in the github repository for this paper (\cite{latent_2020}).

\section{Numerical Demonstration}
\subsection{Synthetic Data}
\begin{figure}[ht!]
  \centering
  \begin{minipage}[t]{0.29\linewidth}
    \includegraphics[width=\linewidth]{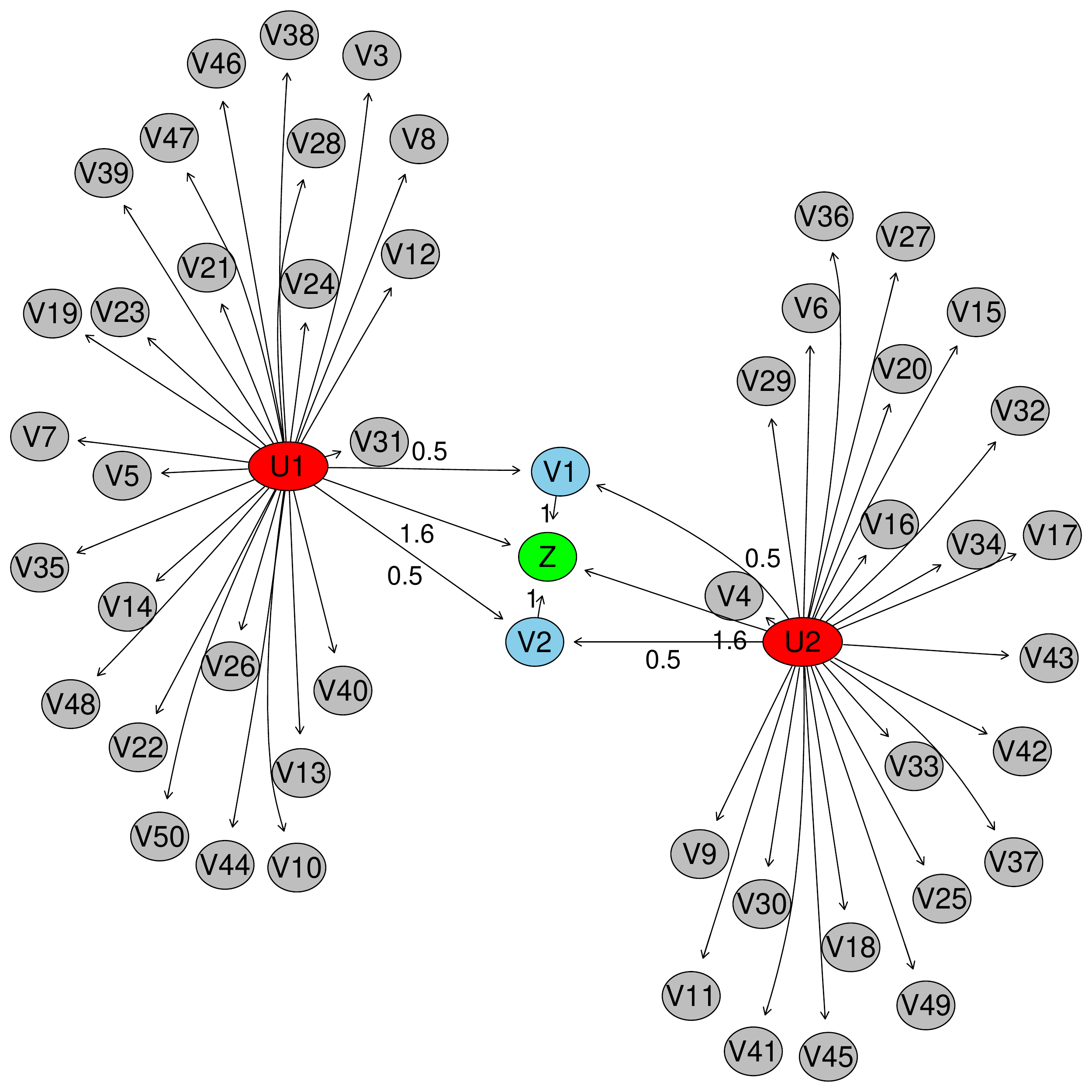}
    \caption{\label{fig_truenet}True network. }
  \end{minipage}\hfill
   \begin{minipage}[t]{0.33\linewidth}
     \includegraphics[width=\linewidth]{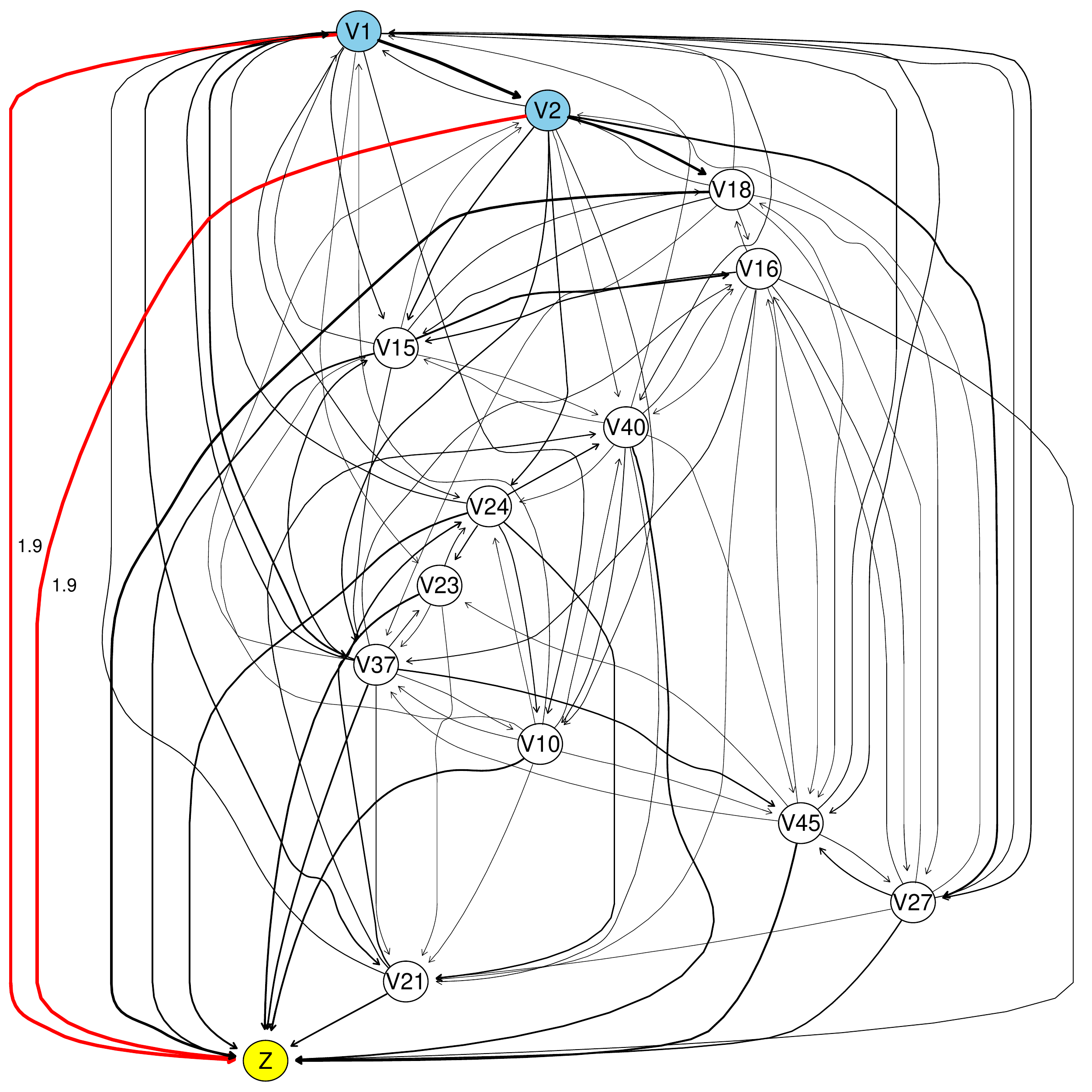}
     \caption{\label{fig_missing} Estimated network when $U_1$ and
      $U_2$ are unobserved.}
   \end{minipage}\hfill
   \begin{minipage}[t]{0.33\linewidth }
    \includegraphics[width=\linewidth]{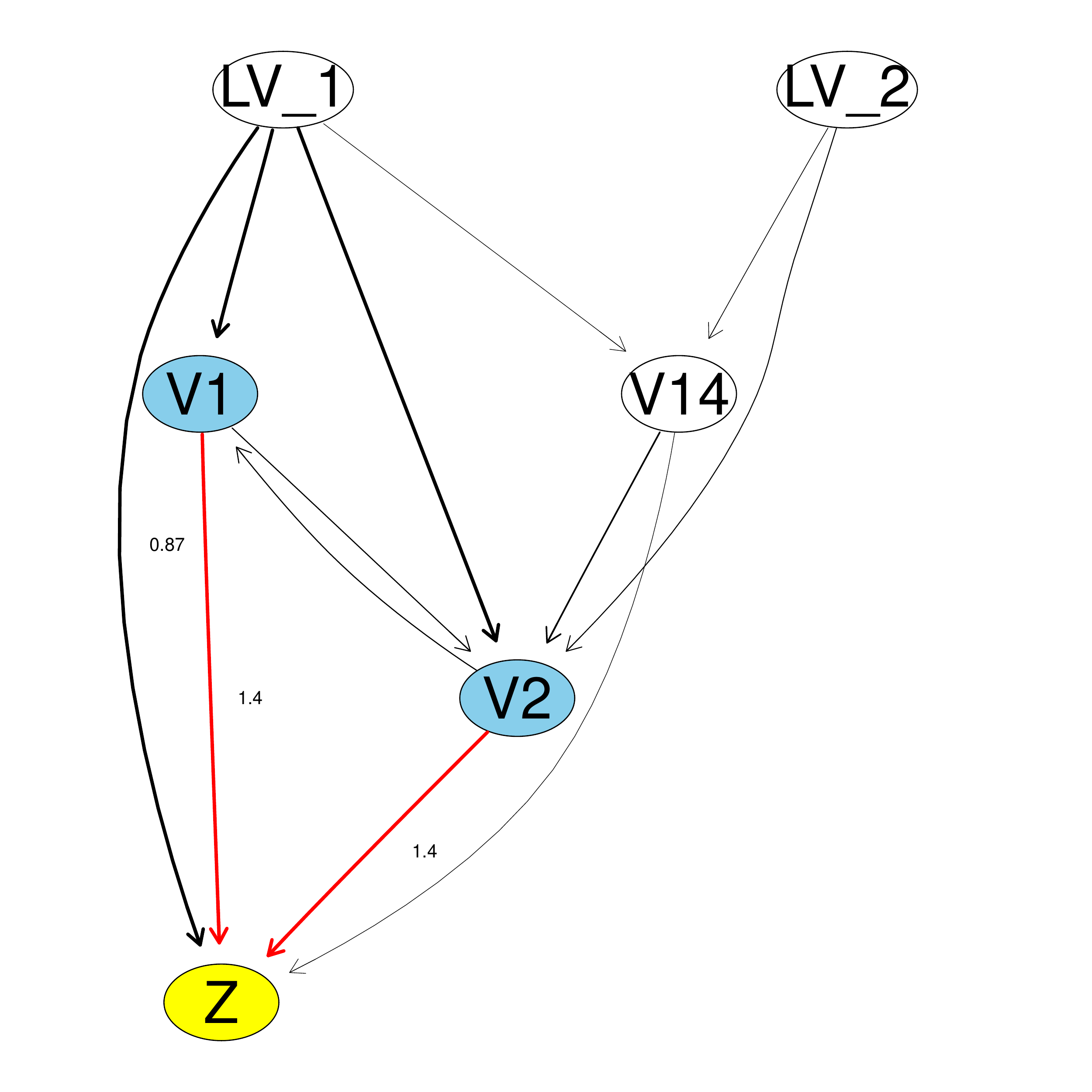}
    \caption{\label{fig_estnet_infered} Estimated network at the last
      iteration of algorithm \ref{alg:latentEM}.}
  \end{minipage}
\end{figure}

To illustrate  the algorithms described in the previous sections we
generated synthetic data from the network shown in Figure
\ref{fig_truenet} (also supplementary Figure \ref{sup-fig_network_true}) where two variables $V_1$ and $V_2$ drive an outcome
$Z$. Two confounders $U_1$ and $U_2$ affect both the drivers and the
outcome, as well as many additional variables that do not affect the
outcome $Z$.
The coefficient values in the network were chosen so that faithfullness (also knows as stability \cite{pearl_causality:_2000}) was achieved allowing the structure and coefficients to be  approximately recovered when all variables were observed (see supplement for additional examples).

The underlying network inference needed for the algorithm was implemented by bootstrapping the data  and running the R package bnlearn (\cite{scutari_learning_2010} also see code in \cite{latent_2020} for specific settings of the run.) on each bootstrap.  The resulting ensemble of networks can be combined to obtain a consensus network where only the most confident edges are kept. Similarly, the coefficient estimates can be obtained by averaging them over bootstraps.

For this example, using the complete data (no missing variables) the consensus network created with edges with confidence larger than 40\% recovers the true structure (see supplementary Figure \ref{sup-fig_network_full}), and the root mean square error (RMSE) in the coefficient estimates is 0.05 (not shown). This represents a lower
bound on the error that we can expect to obtain under perfect reconstruction of the latent space.

When the confounders are unobserved, the reconstruction of the network introduces many false edges and results in a five-times-larger RMSE. Figure \ref{fig_missing} (see also supplementary Figure \ref{sup-fig_network_miss}) shows the reconstructed network, where the red edges are the true edges between $V_1$ and $V_2$ and the outcome $Z$.

\begin{figure}[ht]
  \centering
  \begin{minipage}[t]{0.42\linewidth}
    \includegraphics[scale=0.3]{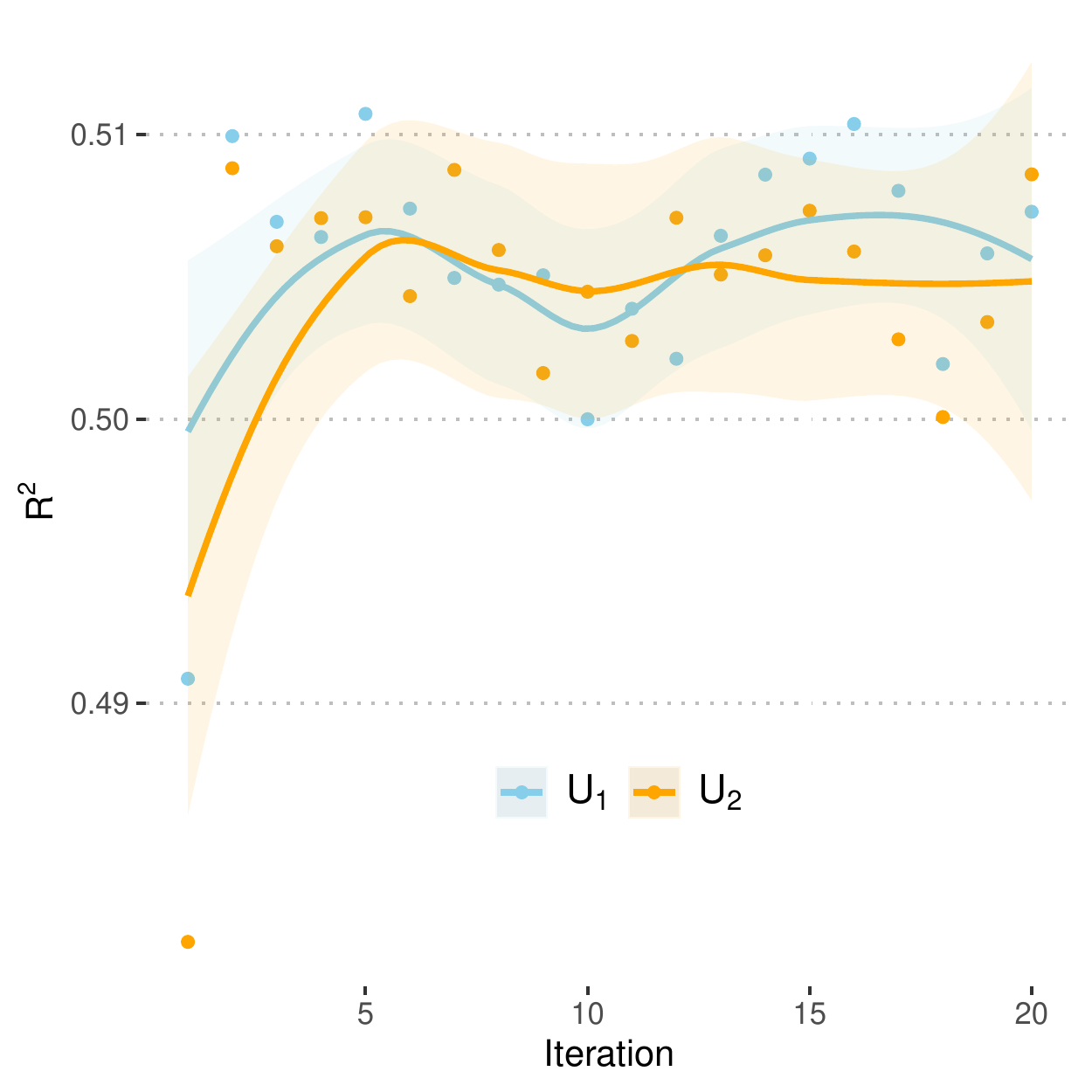}
    \caption{\label{fig:latrecons}$R^2$ in the prediction of the latent variable from the selected principal components.\medskip}
  \end{minipage}\hfill
  \begin{minipage}[t]{0.42\linewidth}
    \includegraphics[scale=0.3]{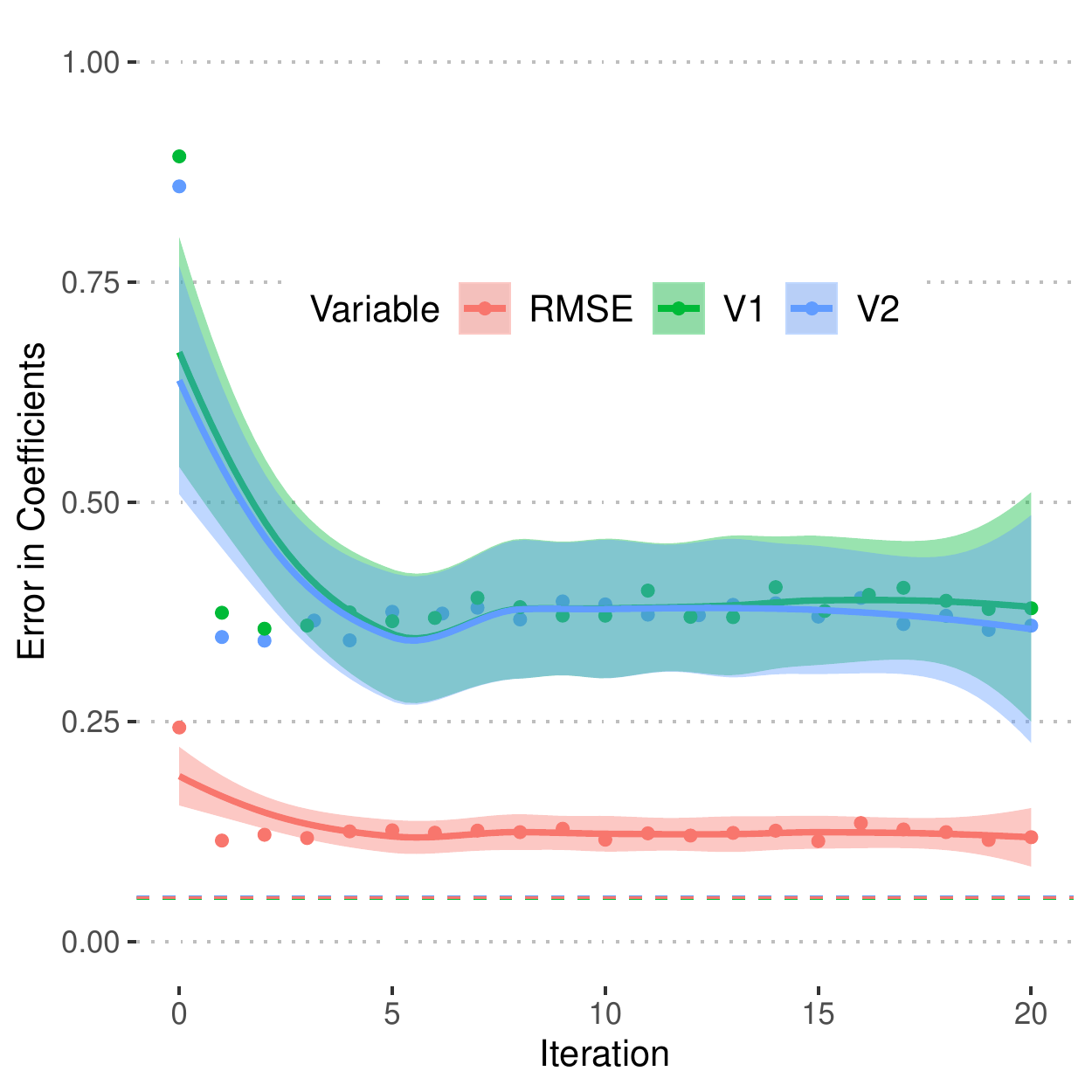}
    \caption{\label{fig:errorcoef}Error in coefficients as a function of the iterations.\medskip}
  \end{minipage}
\end{figure}

We ran \ref{alg:latentEM} for 20 iterations using PCA to reconstruct the latent space from the residuals with the assumption that the latent variables were source nodes. We then tracked the latent variable reconstruction in an out-of-sample test set as well as the error in the coefficient's estimates. Figure \ref{fig:latrecons} (supplementary Figure \ref{sup-fig_latvar_r2}) shows the adjusted $R^2$ between each of the true latent variables and the prediction obtained from the estimated latent space across iterations. The lines and error bands are calculated using locally estimated scatterplot smoothing (LOESS). The estimated latent space is predictive of both the latent variables, and the iterative procedure improves the $R^2$ with respect to $U_1$ from 0.49 at the first iterations to about 0.505 in about 3 iterations.

Figure \ref{fig:errorcoef} (also supplementary Figure \ref{sup-fig_coef_rmse}) shows the total error in the coefficients
of all variables connected to the outcome $Z$ (RMSE) in the infered network as
well as the error in the coefficients of the true drivers of $Z$, $V_1$
and $V_2$. The dashed lines show the error levels when all variables are observed.

Figure \ref{fig_estnet_infered}  (and supplementary Figure 7) shows the final inferred network at iteration 20. The number of edges arriving to the outcome was reduced considerably with respect to the network prior to inferring latent variables (Figure \ref{fig_missing} and supplementary Figure \ref{sup-fig_network_learn}). In addition, the coefficients connecting V1 and V2 to the outcome are now closer to their true values (Figure \ref{fig:errorcoef}). This represents an improvement in ACE estimation.


\subsection{Experimental Data}
In order to assess the efficacy of our approach to latent variable inference on realistic data, we picked a dataset of a type that is typically hard to analyze.  The dataset in question was obtained from the Cure Huntington's Disease Initiative (CHDI) foundation \cite{langfelder_integrated_2016}.  It captures molecular phenotyping and length of Huntingtin gene's CAG repeat expansion in CAG knock-in mice and consists of striatum gene expression and proteomic data as well as other measurements.  For this study, we extracted CAG, mouse weight and age, and a number of gene expression and proteomic measurements correlated with CAG, so that gene expression would also have matching proteomic data, for a total of 249 variables.over 153 samples. 

\begin{figure}[ht!]
  \centering
    \includegraphics[width=\linewidth,page=2]{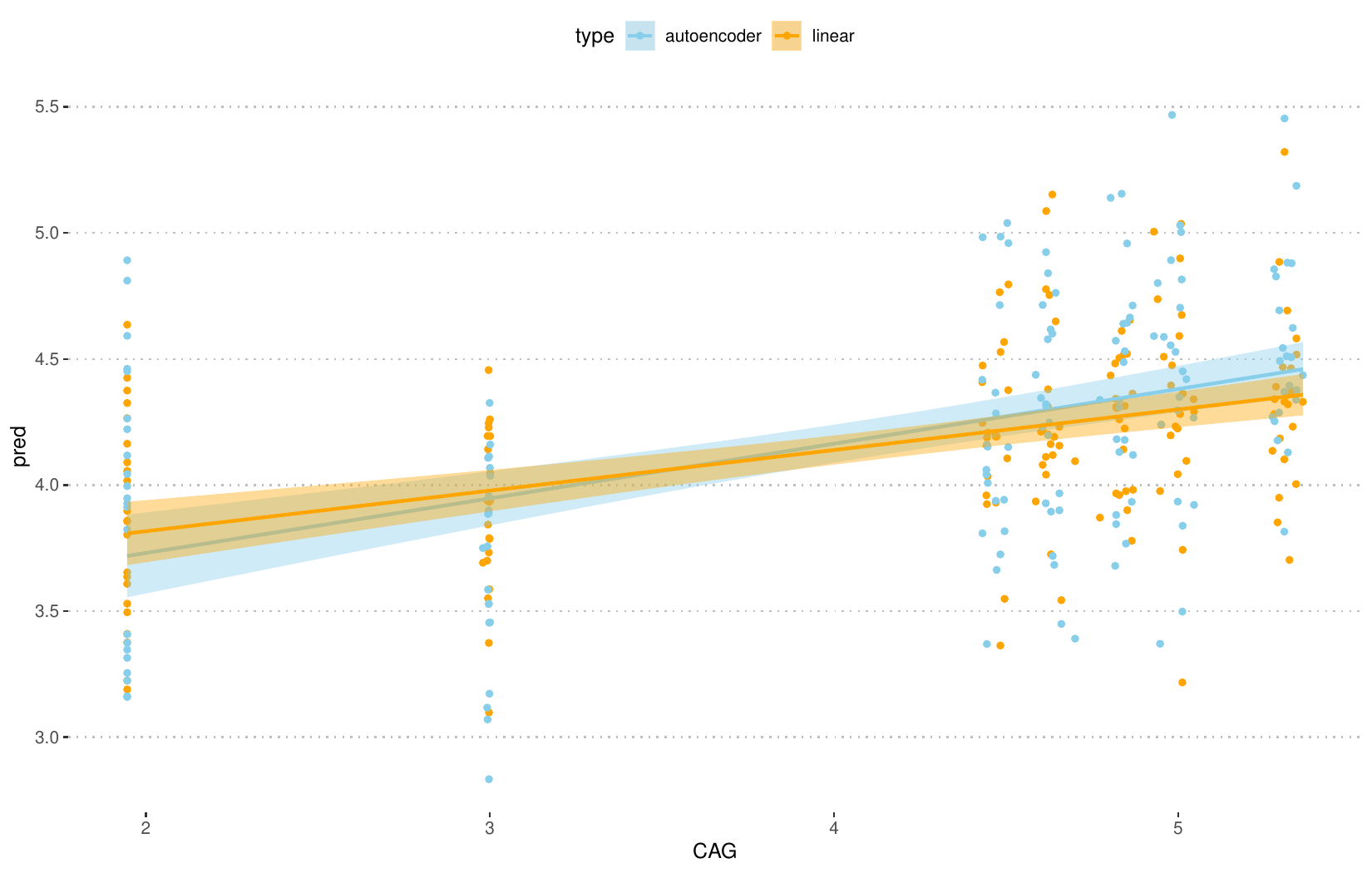}
    \caption{\label{fig_cag_linear} Autoencoder predicts CAG slightly better than linear PCA}
\end{figure}

Excessive CAG repeats cause Huntingont's disease via a complex and not entirely understood mechanism, where disease age of onset and severity strongly correlate with repeat length.  Furthermore, CAG influences a large number of gene expression and protein level markers, based on bayesian modeling - mostly indirectly.  Therefore CAG is an ideal pleiotropic covariate of a type that our algorithm should be able to uncover if it were missing.  However, the high biological noise, low sample size, and the presence of ties and near-ties in CAG values leave enough difficulty in the problem.  Since we didin't have an outcome with a lot of signal on which to test effects of latent confounding, we focused on inference of the latent variable as our measure of success in this test (Supplementary Algorithm \ref{sup-alg_cag}).  We developed an autoencoder approach to inference for cases latent space may relate to observables in a very non-linear manner.  In this case, we begin to see marginally better performance for autoencoder, though on simulated data linear approach works better, as expected (Figure \ref{fig_cag_linear}).  Given the small number of samples, a problem for deep learning, we believe that autoencoder is a promising approach for real-world datasets, especially biological data.

\medskip

\section{Generalizations of this approach}
In this section, we discuss extensions to this approach for GGMs with interactions, nonlinear functional forms, and categorical variables. 

\subsection{Gaussian Graphical Models With Interactions}
In the presence of interactions among variables in a GGM, equation \ref{eq:residualColumn} expressing the deviation of residuals from Gaussian noise may acquire higher-order terms due to interactions among the descendants of the latent space $U$:

\begin{equation}
R_N = \beta_0 + \beta_1 U + \beta_2 U^2 + \beta_2 U^3 + \dots.
\label{eq:residualColumnWithInteractions}
\end{equation}

Assuming interactions up to $k$th power are present in the system being modeled, residuals for each variable may have up to $k$ terms in the model matrix described by equation \ref{eq:residualColumnWithInteractions}, and if interactions among variables in the latent space $U$ also exist, the cardinality of the principal components of the residuals may far exceed the cardinality of the underlying latent space.  Nevertheless, it may be possible to reconstruct a parsimonious basis vector by application of regularization and nonlinear approaches to latent variable modeling, such as nonlinear PCA (e.g. using methods from \cite{karatzoglou_kernlab_2004}), or autoencoders \cite{louizos_causal_2017} as will be discussed below.

\subsection{Generalization to nonlinear functions}
We can show that linear PCA will suffice for a set of models broader than GGMs.  In particular, we will focus on nonlinear but homeomorphic functions within the Generalized Additive Model (GAM) family.  When talking about multiple inputs, we will require that the relationship of any output variable to any of the covariates in equation \ref{eq:residualColumn} is homeomorphic (invertible), and that equation \ref{eq:resPCA} can be marginalized with respect to any right-hand-side variable as well as to the original left-hand side variable.  For such class of transformations, mutual information between variables, such as between a single confounder $U$ and some downstream variable $N$, is invariant \cite{kraskov_estimating_2004}.  Therefore, residuals of any variable $N$ will be rank-correlated to $rank(U)$ in a transformation-invariant way. Further, spearman rank-correlation, specifically, is defined as pearson correlation of ranks, and pearson correlation is a special case of mutual information for bivariate normal distribution.  Therefore when talking about mutual information between ranks of arbitrarily distributed variables, we can use our results for the GGM case above.

Thus, equation \ref{eq:residualColumn} will apply here with some modifications:
\begin{equation}
rank(R_N) = \beta_0 + \beta_1 rank(U) + \xi_N.
\label{eq:residualColumnRank}
\end{equation}

Since a method has been published recently describing how to capture rank-equivalent residuals (aka probability-scale residuals, or PSR) for any ordinal variable \cite{shepherd_probability-scale_2016}, we can modify the equation \ref{eq:resPCA} to reconstruct latent space up to rank-equivalence when interactions are absent from the network.

\begin{equation}
rank(U) = \frac{1}{\beta_i} rank(R_i) + \frac{1}{\beta_j} rank(R_j) + \dots + \xi. 
\label{resPcaGam}
\end{equation}

When $U$ consists of multiple variables that are independent of each other, the relationship between $N$ and $U$ can be written down using the mutual information chain rule \cite{mackay_information_2003} and simplified taking advantage of mutual independence of the latent sources:
\begin{equation}
\label{eq:rankSetRelationship}
\begin{split}
I(N; U) = I(N; U_1, U_2, \dots, U_U) \\= \sum_{i=1}^{u}{I(N; X_i | X_{i-1}, \dots, X_1)} = \sum_{i=1}^{u}{I(N; X_i)}.
\end{split}
\end{equation}

If interactions among $U$ are present, it may still be possible to approximate the latent space with a suitably regularized nonlinear basis, but we do not, at present, know of specific conditions when this may or may not work.  Novel methods for encoding basis sets, such as nonlinear PCA (implemented in the accompanying code), autoencoders, and others, may be brought to bear to collapse the linearly independent basis down to non-linearly independent (i.e. in the mutual information sense) components.

While approximate inference of latent variables for GMs built over invertible functions had been noted in \cite{elidan_ideal_2007}, the above method gives a direct rank-linear approach leveraging the recently-proposed PSRs.  Similar ideas pertaining to the Ideal Parents approach and involving copula transforms have been outlined in \cite{tenzer_generalized_2016}.

\subsection{Generalization to categorical variables}
In principle, PSRs can be extended to the case of non-ordinal categorical variables by modeling binary in/out of class label, deviance being correct/false.  These models would lack the smooth gradient allowed by ranks and would probably converge far worse and offer more local minima for EM to get stuck in.  

\section{Conclusions and Future Directions}
In this work we present a method for describing the latent variable space which is locally optimal under linearity, up to rank-linearity.  The method does not place \textit{a priori} constraints on the number of latent variables, and will infer the upper bound on this dimensionality automatically.  

In real-data situations, some assumptions of our linear approach may not hold - PCA and related methods may not suffice.  For such cases, we provide a heuristic autoencoder implementation.  Autoencoders have shown utility when the structure of the causal model is already known \cite{louizos_causal_2017}.  Learning structure over observed data as well as unconstrained latent space via autoencoders, as in our work, results in a hybrid "deep causal discovery" generalization of that approach.  

Many domains can benefit from joint causal discovery and inference of latent space.  For instance, in epidemiology neither the model structure nor the latent space are typically well-understood beyond simple examples.  Multi-omic biological datasets are another area of applicability, since it's impossible to collect all biological data modalities in practice.  In clinical trials, epidemiological features of multi-omic datasets may be hard to fully account for.  

Combining causal discovery and causal inference improves statements about ACE, and helps assess data set limitations.  Further, we provide a relatively fast implementation suitable to real problems.  Therefore, we hope that our work will open new practical applications of both paradigms.
 
\small
\bibliography{LatentVars-arxiv}

\begin{thebibliography}{38}
\providecommand{\natexlab}[1]{#1}
\providecommand{\url}[1]{\texttt{#1}}
\providecommand{\urlprefix}{URL }
\expandafter\ifx\csname urlstyle\endcsname\relax
  \providecommand{\doi}[1]{doi:\discretionary{}{}{}#1}\else
  \providecommand{\doi}{doi:\discretionary{}{}{}\begingroup
  \urlstyle{rm}\Url}\fi

\bibitem[{Anandkumar et~al.(2013)Anandkumar, Hsu, Javanmard, and
  Kakade}]{anandkumar_learning_2013}
Anandkumar, A.; Hsu, D.; Javanmard, A.; and Kakade, S. 2013.
\newblock Learning {Linear} {Bayesian} {Networks} with {Latent} {Variables}.
\newblock In Dasgupta, S.; and McAllester, D., eds., \emph{Proceedings of the
  30th {International} {Conference} on {Machine} {Learning}}, volume~28 of
  \emph{Proceedings of {Machine} {Learning} {Research}}, 249--257. Atlanta,
  Georgia, USA: PMLR.
\newblock \urlprefix\url{http://proceedings.mlr.press/v28/anandkumar13.html}.

\bibitem[{Bareinboim and Pearl(2016)}]{bareinboim_causal_2016}
Bareinboim, E.; and Pearl, J. 2016.
\newblock Causal inference and the data-fusion problem.
\newblock \emph{Proceedings of the National Academy of Sciences} 113(27):
  7345--7352.
\newblock ISSN 0027-8424, 1091-6490.
\newblock \doi{10.1073/pnas.1510507113}.
\newblock
  \urlprefix\url{http://www.pnas.org/lookup/doi/10.1073/pnas.1510507113}.

\bibitem[{Berlinkov(2018)}]{pca_ae}
Berlinkov, M. 2018.
\newblock python - {Tying} {Autoencoder} {Weights} in a {Dense} {Keras}
  {Layer}.
\newblock
  \urlprefix\url{https://stackoverflow.com/questions/53751024/tying-autoencoder-weights-in-a-dense-keras-layer}.

\bibitem[{Bernstein et~al.(2020)Bernstein, Saeed, Squires, and
  Uhler}]{bernstein2020ordering}
Bernstein, D.; Saeed, B.; Squires, C.; and Uhler, C. 2020.
\newblock Ordering-Based Causal Structure Learning in the Presence of Latent
  Variables.
\newblock In \emph{International Conference on Artificial Intelligence and
  Statistics}, 4098--4108. {PMLR}.

\bibitem[{Buja and Eyuboglu(1992)}]{buja_remarks_1992}
Buja, A.; and Eyuboglu, N. 1992.
\newblock Remarks on {Parallel} {Analysis}.
\newblock \emph{Multivariate Behavioral Research} 27(4): 509--540.
\newblock ISSN 0027-3171, 1532-7906.
\newblock \doi{10.1207/s15327906mbr2704_2}.
\newblock
  \urlprefix\url{http://www.tandfonline.com/doi/abs/10.1207/s15327906mbr2704_2}.

\bibitem[{Claassen, Mooij, and Heskes(2013)}]{claassenLearningSparseCausal2013}
Claassen, T.; Mooij, J.; and Heskes, T. 2013.
\newblock Learning {{Sparse Causal Models}} Is Not {{NP}}-Hard.
\newblock In \emph{Proceedings of the {{Twenty}}-{{Ninth Conference}} on
  {{Uncertainty}} in {{Artificial Intelligence}}}, 172--181.
\newblock \urlprefix\url{http://arxiv.org/abs/1309.6824}.

\bibitem[{Colombo et~al.(2012)Colombo, Maathuis, Kalisch, and
  Richardson}]{colombo_learning_2012}
Colombo, D.; Maathuis, M.~H.; Kalisch, M.; and Richardson, T.~S. 2012.
\newblock Learning High-Dimensional Directed Acyclic Graphs with Latent and
  Selection Variables.
\newblock \emph{The Annals of Statistics} 40(1): 294--321.
\newblock ISSN 0090-5364.
\newblock \doi{10.1214/11-AOS940}.

\bibitem[{D'Amour(2019)}]{damour_multi-cause_2019}
D'Amour, A. 2019.
\newblock On {Multi}-{Cause} {Causal} {Inference} with {Unobserved}
  {Confounding}: {Counterexamples}, {Impossibility}, and {Alternatives}.
\newblock \emph{arXiv:1902.10286 [cs, stat]}
  \urlprefix\url{http://arxiv.org/abs/1902.10286}.
\newblock ArXiv: 1902.10286.

\bibitem[{Elidan et~al.(2001)Elidan, Lotner, Friedman, and
  Koller}]{elidan_discovering_2001}
Elidan, G.; Lotner, N.; Friedman, N.; and Koller, D. 2001.
\newblock Discovering {Hidden} {Variables}: {A} {Structure}-{Based} {Approach}.
\newblock In Leen, T.~K.; Dietterich, T.~G.; and Tresp, V., eds.,
  \emph{Advances in {Neural} {Information} {Processing} {Systems} 13},
  479--485. MIT Press.
\newblock
  \urlprefix\url{http://papers.nips.cc/paper/1940-discovering-hidden-variables-a-structure-based-approach.pdf}.

\bibitem[{Elidan, Nachman, and Friedman(2007)}]{elidan_ideal_2007}
Elidan, G.; Nachman, I.; and Friedman, N. 2007.
\newblock “{Ideal} {Parent}” {Structure} {Learning} for {Continuous}
  {Variable} {Bayesian} {Networks}.
\newblock \emph{Journal of Machine Learning Research} 8: 35.

\bibitem[{Friedman(1997)}]{friedman1997learning}
Friedman, N. 1997.
\newblock Learning belief networks in the presence of missing values and hidden
  variables.
\newblock In \emph{{ICML}}, volume~97, 125--133.

\bibitem[{Friedman(1998)}]{friedman1998bayesian}
Friedman, N. 1998.
\newblock The {Bayesian} structural {EM} algorithm.
\newblock In \emph{Proceedings of the {Fourteenth} conference on {Uncertainty}
  in artificial intelligence}, 129--138.

\bibitem[{Friedman and Koller(2013)}]{friedman_being_2013}
Friedman, N.; and Koller, D. 2013.
\newblock Being {Bayesian} about {Network} {Structure}.
\newblock \emph{arXiv:1301.3856 [cs, stat]}
  \urlprefix\url{http://arxiv.org/abs/1301.3856}.
\newblock ArXiv: 1301.3856.

\bibitem[{Frot, Nandy, and Maathuis(2017)}]{frotRobustCausalStructure2017}
Frot, B.; Nandy, P.; and Maathuis, M.~H. 2017.
\newblock Robust Causal Structure Learning with Some Hidden Variables.
\newblock \emph{arXiv:1708.01151 [stat]} .

\bibitem[{Gavish and Donoho(2014)}]{gavish_optimal_2014}
Gavish, M.; and Donoho, D.~L. 2014.
\newblock The {Optimal} {Hard} {Threshold} for {Singular} {Values} is
  {\textbackslash}(4/{\textbackslash}sqrt \{3\}{\textbackslash}).
\newblock \emph{IEEE Transactions on Information Theory} 60(8): 5040--5053.
\newblock ISSN 0018-9448, 1557-9654.
\newblock \doi{10.1109/TIT.2014.2323359}.
\newblock \urlprefix\url{http://ieeexplore.ieee.org/document/6846297/}.

\bibitem[{github(2020)}]{latent_2020}
github. 2020.
\newblock Latent {Confounder}.
\newblock \urlprefix\url{https://github.com/rimorob/netres}.
\newblock Original-date: 2020-02-05T03:35:01Z.

\bibitem[{Hernán and Robins(2006)}]{hernan_estimating_2006}
Hernán, M.~A.; and Robins, J.~M. 2006.
\newblock Estimating causal effects from epidemiological data.
\newblock \emph{Journal of Epidemiology and Community Health} 60(7): 578--586.
\newblock ISSN 0143-005X.
\newblock \doi{10.1136/jech.2004.029496}.

\bibitem[{Ioffe and Szegedy(2015)}]{ioffe2015batch}
Ioffe, S.; and Szegedy, C. 2015.
\newblock Batch Normalization: Accelerating Deep Network Training by Reducing
  Internal Covariate Shift.

\bibitem[{Jabbari et~al.(2017)Jabbari, Ramsey, Spirtes, and
  Cooper}]{jabbariDiscoveryCausalModels2017}
Jabbari, F.; Ramsey, J.; Spirtes, P.; and Cooper, G. 2017.
\newblock Discovery of {{Causal Models}} That {{Contain Latent Variables}}
  through {{Bayesian Scoring}} of {{Independence Constraints}}.
\newblock \emph{Joint European Conference on Machine Learning and Knowledge
  Discovery in Databases} 142--157.
\newblock \doi{10.1007/978-3-319-71246-8_9}.

\bibitem[{Karatzoglou et~al.(2004)Karatzoglou, Smola, Hornik, and
  Zeileis}]{karatzoglou_kernlab_2004}
Karatzoglou, A.; Smola, A.; Hornik, K.; and Zeileis, A. 2004.
\newblock \textbf{kernlab} - {An} \textit{{S}4} {Package} for {Kernel}
  {Methods} in \textit{{R}}.
\newblock \emph{Journal of Statistical Software} 11(9).
\newblock ISSN 1548-7660.
\newblock \doi{10.18637/jss.v011.i09}.
\newblock \urlprefix\url{http://www.jstatsoft.org/v11/i09/}.

\bibitem[{Koller and Friedman(2009)}]{koller_probabilistic_2009}
Koller, D.; and Friedman, N. 2009.
\newblock \emph{Probabilistic graphical models: principles and techniques}.
\newblock Adaptive computation and machine learning. Cambridge, MA: MIT Press.
\newblock ISBN 978-0-262-01319-2.

\bibitem[{Kraskov, Stögbauer, and Grassberger(2004)}]{kraskov_estimating_2004}
Kraskov, A.; Stögbauer, H.; and Grassberger, P. 2004.
\newblock Estimating mutual information.
\newblock \emph{Physical Review E} 69(6).
\newblock ISSN 1539-3755, 1550-2376.
\newblock \doi{10.1103/PhysRevE.69.066138}.
\newblock \urlprefix\url{https://link.aps.org/doi/10.1103/PhysRevE.69.066138}.

\bibitem[{Langfelder et~al.(2016)Langfelder, Cantle, Chatzopoulou, Wang, Gao,
  Al-Ramahi, Lu, Ramos, El-Zein, Zhao, Deverasetty, Tebbe, Schaab, Lavery,
  Howland, Kwak, Botas, Aaronson, Rosinski, Coppola, Horvath, and
  Yang}]{langfelder_integrated_2016}
Langfelder, P.; Cantle, J.~P.; Chatzopoulou, D.; Wang, N.; Gao, F.; Al-Ramahi,
  I.; Lu, X.-H.; Ramos, E.~M.; El-Zein, K.; Zhao, Y.; Deverasetty, S.; Tebbe,
  A.; Schaab, C.; Lavery, D.~J.; Howland, D.; Kwak, S.; Botas, J.; Aaronson,
  J.~S.; Rosinski, J.; Coppola, G.; Horvath, S.; and Yang, X.~W. 2016.
\newblock Integrated genomics and proteomics to define huntingtin {CAG}
  length-dependent networks in {HD} {Mice}.
\newblock \emph{Nature neuroscience} 19(4): 623--633.
\newblock ISSN 1097-6256.
\newblock \doi{10.1038/nn.4256}.
\newblock
  \urlprefix\url{https://www.ncbi.nlm.nih.gov/pmc/articles/PMC5984042/}.

\bibitem[{Louizos et~al.(2017)Louizos, Shalit, Mooij, Sontag, Zemel, and
  Welling}]{louizos_causal_2017}
Louizos, C.; Shalit, U.; Mooij, J.; Sontag, D.; Zemel, R.; and Welling, M.
  2017.
\newblock Causal {Effect} {Inference} with {Deep} {Latent}-{Variable} {Models}.
\newblock \emph{arXiv:1705.08821 [cs, stat]}
  \urlprefix\url{http://arxiv.org/abs/1705.08821}.
\newblock ArXiv: 1705.08821.

\bibitem[{MacKay(2003)}]{mackay_information_2003}
MacKay, D. J.~C. 2003.
\newblock \emph{Information theory, inference, and learning algorithms}.
\newblock Cambridge, UK ; New York: Cambridge University Press.
\newblock ISBN 978-0-521-64298-9.

\bibitem[{Nandy et~al.(2018)Nandy, Hauser, Maathuis et~al.}]{nandy2018high}
Nandy, P.; Hauser, A.; Maathuis, M.~H.; et~al. 2018.
\newblock High-Dimensional Consistency in Score-Based and Hybrid Structure
  Learning.
\newblock \emph{The Annals of Statistics} 46(6A): 3151--3183.

\bibitem[{Pearl(2000)}]{pearl_causality:_2000}
Pearl, J. 2000.
\newblock \emph{Causality: models, reasoning, and inference}.
\newblock Cambridge, U.K.; New York: Cambridge University Press.
\newblock ISBN 978-1-139-64936-0 978-0-511-80316-1.
\newblock \urlprefix\url{http://dx.doi.org/10.1017/CBO9780511803161}.
\newblock OCLC: 834142635.

\bibitem[{Ranjan(2019)}]{ranjan_build_2019}
Ranjan, C. 2019.
\newblock Build the right {Autoencoder} — {Tune} and {Optimize} using {PCA}
  principles. {Part} {II}.
\newblock
  \urlprefix\url{https://towardsdatascience.com/build-the-right-autoencoder-tune-and-optimize-using-pca-principles-part-ii-24b9cca69bd6}.

\bibitem[{Rawlings, Pantula, and Dickey(1998)}]{rawlings_applied_1998}
Rawlings, J.~O.; Pantula, S.~G.; and Dickey, D.~A. 1998.
\newblock \emph{Applied regression analysis: a research tool}.
\newblock Springer texts in statistics. New York: Springer, 2nd ed edition.
\newblock ISBN 978-0-387-98454-4.

\bibitem[{Scutari(2010)}]{scutari_learning_2010}
Scutari, M. 2010.
\newblock Learning {Bayesian} {Networks} with the \textbf{bnlearn} \textit{{R}}
  {Package}.
\newblock \emph{Journal of Statistical Software} 35(3).
\newblock ISSN 1548-7660.
\newblock \doi{10.18637/jss.v035.i03}.
\newblock \urlprefix\url{http://www.jstatsoft.org/v35/i03/}.

\bibitem[{Shepherd, Li, and Liu(2016)}]{shepherd_probability-scale_2016}
Shepherd, B.~E.; Li, C.; and Liu, Q. 2016.
\newblock Probability-scale residuals for continuous, discrete, and censored
  data.
\newblock \emph{The Canadian journal of statistics = Revue canadienne de
  statistique} 44(4): 463--479.
\newblock ISSN 0319-5724.
\newblock \doi{10.1002/cjs.11302}.
\newblock
  \urlprefix\url{https://www.ncbi.nlm.nih.gov/pmc/articles/PMC5364820/}.

\bibitem[{Silva et~al.(2006)Silva, Scheines, Glymour, and
  Spirtes}]{silva_learning_2006}
Silva, R.; Scheines, R.; Glymour, C.; and Spirtes, P. 2006.
\newblock Learning the {Structure} of {Linear} {Latent} {Variable} {Models}.
\newblock \emph{J. Mach. Learn. Res.} 7: 191--246.

\bibitem[{Spirtes, Glymour, and Scheines(1993)}]{spirtes_causation_1993}
Spirtes, P.; Glymour, C.~N.; and Scheines, R. 1993.
\newblock \emph{Causation, prediction, and search}.
\newblock Number~81 in Lecture notes in statistics. New York: Springer-Verlag.
\newblock ISBN 978-0-387-97979-3.

\bibitem[{Spirtes, Meek, and
  Richardson(1995)}]{spirtesCausalInferencePresence1995}
Spirtes, P.~L.; Meek, C.; and Richardson, T.~S. 1995.
\newblock Causal {{Inference}} in the {{Presence}} of {{Latent Variables}} and
  {{Selection Bias}}.
\newblock In \emph{Proceedings of the {{Eleventh}} Conference on
  {{Uncertainty}} in Artificial Intelligence}, 499--506.
\newblock \urlprefix\url{http://arxiv.org/abs/1302.4983}.

\bibitem[{Tenzer and Elidan(2016)}]{tenzer_generalized_2016}
Tenzer, Y.; and Elidan, G. 2016.
\newblock Generalized {Ideal} {Parent} ({GIP}): {Discovering} non-{Gaussian}
  {Hidden} {Variables}.
\newblock In Gretton, A.; and Robert, C.~C., eds., \emph{Proceedings of
  {Machine} {Learning} {Research}}, volume~51, 222--230. Proceedings of Machine
  Learning Research: PMLR.
\newblock \urlprefix\url{http://proceedings.mlr.press}.

\bibitem[{Ushey(2020)}]{reticulate_2020}
Ushey, K. 2020.
\newblock rstudio/reticulate.
\newblock \urlprefix\url{https://github.com/rstudio/reticulate}.
\newblock Original-date: 2017-02-06T18:59:46Z.

\bibitem[{Wang and Blei(2019{\natexlab{a}})}]{wang_deconfounder_2019}
Wang, Y.; and Blei, D.~M. 2019{\natexlab{a}}.
\newblock The {{Blessings}} of {{Multiple Causes}}.
\newblock \emph{Journal of the American Statistical Association} 114(528):
  1574--1596.

\bibitem[{Wang and Blei(2019{\natexlab{b}})}]{wangMultipleCausesCausal2019}
Wang, Y.; and Blei, D.~M. 2019{\natexlab{b}}.
\newblock Multiple {{Causes}}: {{A Causal Graphical View}}.
\newblock \emph{arXiv:1905.12793 [cs, stat]} .

\end{thebibliography}

\end{document}